\documentclass{article}

\usepackage{microtype}
\usepackage{graphicx}
\usepackage{subcaption}
\usepackage{booktabs} 

\usepackage{hyperref}

\usepackage{rotating}
\usepackage{tikz}
\usetikzlibrary{
  arrows.meta,
  positioning,
  calc,
  fit,
  backgrounds
}
\usepackage{dblfloatfix}

\usepackage[preprint]{icml2026_fogen}

\usepackage{amsmath}
\usepackage{amssymb}
\usepackage{mathtools}
\usepackage{amsthm}

\usepackage[capitalize,noabbrev]{cleveref}

\theoremstyle{plain}

\theoremstyle{definition}

\theoremstyle{remark}

\usepackage[textsize=tiny]{todonotes}

\icmltitlerunning{Generalization by Permutation Routing}

\begin{document}

\twocolumn[
  \icmltitle{Improving Generalization by Permutation Routing Across Model Copies}



  \icmlsetsymbol{equal}{*}

  \begin{icmlauthorlist}
    \icmlauthor{Shuhei Kashiwamura}{equal,yyy}
    \icmlauthor{Timothee Leleu}{equal,yyy,comp}
  \end{icmlauthorlist}

  \icmlaffiliation{yyy}{NTT Research, CA, USA}
  \icmlaffiliation{comp}{Stanford University, CA, USA}

  \icmlcorrespondingauthor{Shuhei Kashiwamura}{shuhei.kashiwamura@ntt-research.com}
  \icmlcorrespondingauthor{Timothee Leleu}{timothee.leleu@ntt-research.com,tleleu@stanford.edu}

  \icmlkeywords{Machine Learning, ICML}

  \vskip 0.3in
]



\printAffiliationsAndNotice{}  

\begin{abstract}
We introduce a use of the \(M\)-cover (or \(M\)-layer) transform for machine
learning.  The method replicates a model \(M\) times, but instead of coupling
the copies through parameter averaging or an explicit attractive force, as in replicated SGD or Elastic SGD, it rewires the contexts in which local learning messages are computed.  Each local loss is evaluated on a routed model whose parameters are drawn from different copies according to permutations sampled from a structured mixing kernel \(Q\).  Training then uses the original local update rule, while the resulting learning messages are redistributed across the copies through these routed computational paths. Thus \(Q\) defines a topology for message transport and controls the long-loop structure of the lifted factor graph. We formulate this construction for perceptrons, committee machines, and multilayer perceptrons, showing that the same principle applies from discrete models to differentiable neural networks.  The resulting framework provides a mechanism for improving generalization through structured message sharing rather than replica collapse or parameter-space coupling.
\end{abstract}

\section{Introduction}
Learning dynamics can become challenging in discrete, quantized, or weakly differentiable models, where the resulting objectives are often non-smooth, piecewise-constant, or highly rugged \cite{yin2019understanding, bengio2013estimating,helwegen2019latent}. Depending on the regime, such systems may exhibit poor local minima, saddles, plateaus, or isolated solutions, leading to slow convergence \cite{dauphin2014identifying, ichikawa2025high} and unstable generalization behavior \cite{keskar2016large, hochreiter1997flat, baldassi2022learning, kashiwamura2024effect}. As a result, optimization remains a central challenge in modern machine learning.

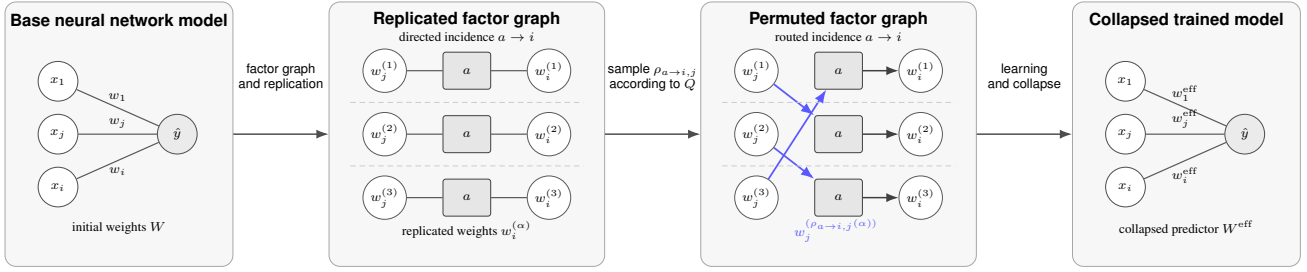
\begin{figure*}[!t]
\centering
\resizebox{\textwidth}{!}{%
\begin{tikzpicture}[
  font=\sffamily,
  >=Latex,
  block/.style={
    rounded corners=6pt,
    draw=black!35,
    fill=black!3,
    minimum width=4.3cm,
    minimum height=5.0cm,
    inner sep=8pt
  },
  title/.style={
    font=\sffamily\bfseries\small,
    align=center
  },
  trans/.style={
    font=\sffamily\scriptsize,
    align=center
  },
  neuron/.style={
    circle,
    draw=black!70,
    fill=white,
    minimum size=7.5mm,
    inner sep=0pt,
    font=\scriptsize
  },
  var/.style={
    circle,
    draw=black!65,
    fill=white,
    minimum size=7.0mm,
    inner sep=0pt,
    font=\scriptsize
  },
  factor/.style={
    rectangle,
    draw=black!75,
    fill=black!10,
    rounded corners=1.5pt,
    minimum width=9mm,
    minimum height=7mm,
    inner sep=1pt,
    font=\scriptsize
  },
  covernode/.style={
    circle,
    draw=black!65,
    fill=white,
    minimum size=8.2mm,
    inner sep=0pt,
    font=\scriptsize
  },
  edge/.style={
    line width=0.65pt,
    draw=black!65
  },
  mainarrow/.style={
    ->,
    line width=0.8pt,
    draw=black!70
  },
  routeedge/.style={
    ->,
    line width=0.95pt,
    draw=blue!65
  },
  fixededge/.style={
    ->,
    line width=0.8pt,
    draw=black!75
  }
]


\node[block] (B1) at (0,0) {};
\node[block, minimum width=5.2cm, right=1.8cm of B1] (B2) {};
\node[block, minimum width=5.2cm, right=1.8cm of B2] (B3) {};
\node[block, right=1.8cm of B3] (B4) {};

\node[title] at ($(B1.north)+(0,-0.35)$) {Base neural network model};
\node[title] at ($(B2.north)+(0,-0.35)$) {Replicated factor graph};
\node[title] at ($(B3.north)+(0,-0.35)$) {Permuted factor graph};
\node[title] at ($(B4.north)+(0,-0.35)$) {Collapsed trained model};

\draw[mainarrow] (B1.east) -- (B2.west);
\draw[mainarrow] (B2.east) -- (B3.west);
\draw[mainarrow] (B3.east) -- (B4.west);

\node[trans] at ($ (B1.east)!0.5!(B2.west) + (0,1.08) $)
{factor graph\\and replication};

\node[trans] at ($ (B2.east)!0.5!(B3.west) + (0,1.08) $)
{sample $\rho_{a\to i,j}$\\according to $Q$};

\node[trans] at ($ (B3.east)!0.5!(B4.west) + (0,1.08) $)
{learning\\and collapse};


\coordinate (b1c) at (B1.center);

\node[neuron] (x1) at ($(b1c)+(-1.15,1.00)$) {$x_1$};
\node[neuron] (x2) at ($(b1c)+(-1.15,0.00)$) {$x_j$};
\node[neuron] (x3) at ($(b1c)+(-1.15,-1.00)$) {$x_i$};
\node[neuron, fill=black!8] (y1) at ($(b1c)+(1.10,0)$) {$\hat y$};

\draw[edge] (x1) -- node[midway, above, font=\scriptsize] {$w_1$} (y1);
\draw[edge] (x2) -- node[midway, above, font=\scriptsize] {$w_j$} (y1);
\draw[edge] (x3) -- node[midway, below, font=\scriptsize] {$w_i$} (y1);

\node[font=\scriptsize] at ($(b1c)+(0,-1.78)$) {initial weights $W$};


\coordinate (b2c) at (B2.center);

\node[covernode] (s1) at ($(b2c)+(-1.55,1.20)$) {$w_j^{(1)}$};
\node[factor]    (f1) at ($(b2c)+(0.00,1.20)$) {$a$};
\node[covernode] (d1) at ($(b2c)+(1.55,1.20)$) {$w_i^{(1)}$};

\node[covernode] (s2) at ($(b2c)+(-1.55,0.00)$) {$w_j^{(2)}$};
\node[factor]    (f2) at ($(b2c)+(0.00,0.00)$) {$a$};
\node[covernode] (d2) at ($(b2c)+(1.55,0.00)$) {$w_i^{(2)}$};

\node[covernode] (s3) at ($(b2c)+(-1.55,-1.20)$) {$w_j^{(3)}$};
\node[factor]    (f3) at ($(b2c)+(0.00,-1.20)$) {$a$};
\node[covernode] (d3) at ($(b2c)+(1.55,-1.20)$) {$w_i^{(3)}$};

\draw[edge] (s1) -- (f1);
\draw[edge] (f1) -- (d1);

\draw[edge] (s2) -- (f2);
\draw[edge] (f2) -- (d2);

\draw[edge] (s3) -- (f3);
\draw[edge] (f3) -- (d3);

\node[font=\scriptsize] at ($(b2c)+(0,1.85)$) {directed incidence $a \to i$};

\draw[densely dashed, draw=black!20] ($(b2c)+(-2.2,0.60)$) -- ($(b2c)+(2.2,0.60)$);
\draw[densely dashed, draw=black!20] ($(b2c)+(-2.2,-0.60)$) -- ($(b2c)+(2.2,-0.60)$);

\node[font=\scriptsize] at ($(b2c)+(0,-1.82)$) {replicated weights $w_i^{(\alpha)}$};


\coordinate (b3c) at (B3.center);

\node[covernode] (ps1) at ($(b3c)+(-1.55,1.20)$) {$w_j^{(1)}$};
\node[factor]    (pf1) at ($(b3c)+(0.00,1.20)$) {$a$};
\node[covernode] (pd1) at ($(b3c)+(1.55,1.20)$) {$w_i^{(1)}$};

\node[covernode] (ps2) at ($(b3c)+(-1.55,0.00)$) {$w_j^{(2)}$};
\node[factor]    (pf2) at ($(b3c)+(0.00,0.00)$) {$a$};
\node[covernode] (pd2) at ($(b3c)+(1.55,0.00)$) {$w_i^{(2)}$};

\node[covernode] (ps3) at ($(b3c)+(-1.55,-1.20)$) {$w_j^{(3)}$};
\node[factor]    (pf3) at ($(b3c)+(0.00,-1.20)$) {$a$};
\node[covernode] (pd3) at ($(b3c)+(1.55,-1.20)$) {$w_i^{(3)}$};

\draw[routeedge] (ps1) -- (pf2);
\draw[routeedge] (ps2) -- (pf3);
\draw[routeedge] (ps3) -- (pf1);

\draw[fixededge] (pf1) -- (pd1);
\draw[fixededge] (pf2) -- (pd2);
\draw[fixededge] (pf3) -- (pd3);

\node[font=\scriptsize] at ($(b3c)+(0,1.85)$) {routed incidence $a \to i$};
\node[font=\scriptsize, text=blue!65] at ($(b3c)+(-0.05,-1.82)$)
{$w_j^{(\rho_{a\to i,j}(\alpha))}$};

\draw[densely dashed, draw=black!20] ($(b3c)+(-2.2,0.60)$) -- ($(b3c)+(2.2,0.60)$);
\draw[densely dashed, draw=black!20] ($(b3c)+(-2.2,-0.60)$) -- ($(b3c)+(2.2,-0.60)$);


\coordinate (b4c) at (B4.center);

\node[neuron] (tx1) at ($(b4c)+(-1.15,1.00)$) {$x_1$};
\node[neuron] (tx2) at ($(b4c)+(-1.15,0.00)$) {$x_j$};
\node[neuron] (tx3) at ($(b4c)+(-1.15,-1.00)$) {$x_i$};
\node[neuron, fill=black!8] (ty1) at ($(b4c)+(1.10,0)$) {$\hat y$};

\draw[edge] (tx1) -- node[midway, above, font=\scriptsize] {$w_1^{\mathrm{eff}}$} (ty1);
\draw[edge] (tx2) -- node[midway, above, font=\scriptsize] {$w_j^{\mathrm{eff}}$} (ty1);
\draw[edge] (tx3) -- node[midway, below, font=\scriptsize] {$w_i^{\mathrm{eff}}$} (ty1);

\node[font=\scriptsize] at ($(b4c)+(0,-1.78)$) {collapsed predictor $W^{\mathrm{eff}}$};

\end{tikzpicture}%
}
\caption{
Workflow of the \(M\)-layer transform for learning.
A base neural-network model with initial weights \(W\) is first interpreted as a factor graph and replicated into \(M\) cover copies.
For a directed incidence \(a\to i\), source variables such as \(w_j\) provide the computational context for the destination variable \(w_i\).
The \(M\)-layer transform preserves the local factor \(a\) but permutes source cover indices according to \(\rho_{a\to i,j}\sim\mathbb P_Q\), so that destination copy \(w_i^{(\alpha)}\) receives context from routed source copies \(w_j^{(\rho_{a\to i,j}(\alpha))}\).
Training is performed on the routed factor graph using the original local update rule, after which the trained cover copies are collapsed into a single effective predictor \(W^{\mathrm{eff}}\) with the same architecture as the base model.
}
\label{fig:mlayer_workflow}
\end{figure*}

A useful perspective comes from graphical models \cite{jordan2004graphical, bishop2006pattern}, where dependencies among variables are represented through graph connectivity. In such systems, local interactions determine which variables directly constrain one another, while the global connectivity structure gives rise to loops and long-range correlations that shape optimization difficulty \cite{weiss2000correctness, mezard2009information}.

Graph-cover constructions provide a principled way to modify global loop structure while preserving local interactions~\cite{vontobel2013counting, altieri2017loop, angelini2024bethe, leleu2026reshaping}. In our framework, we consider an \(M\)-cover of the learning factor graph, where each trainable variable \(w_i\) is replicated into \(M\) cover copies
$
w_i^{(\alpha)},
\quad
\alpha=1,\dots,M.
$
All covers share the same training dataset \(D\), while possessing independent parameter realizations. When the covers are trained independently, local interactions involve only variables carrying the same cover index \(\alpha\). In contrast, the \(M\)-cover construction allows local computations to combine trainable variables originating from different covers through routing permutations. Schematically, a local interaction that originally depends on
$
(w_i,w_j,\ldots)
$
is replaced by a routed interaction of the form
$
(w_i^{(\alpha)}, w_j^{(\beta)}, \ldots),
$
where the cover indices are reassigned according to structured permutations, as illustrated in Figure~\ref{fig:mlayer_workflow}. In this way, the local functional form of the interaction is preserved, while the global connectivity structure through which information propagates is modified.

From the viewpoint of graphical models, the \(M\)-cover therefore reconnects dependency relations among replicated variables without changing the underlying local learning rule. The topology of these routed interactions is controlled by a mixing kernel
$
Q\in\mathbb R_+^{M\times M},
$
which determines how information is transported across the cover dimension.

The probability that cover index $\beta$ is routed into cover $\alpha$ is controlled by $Q_{\alpha\beta}$ which specifies the probability or strength of cross-cover reconnection. In this way, the local computational rule itself remains unchanged, while the global topology through which gradients, errors, or learning signals propagate across the replicated system is explicitly controlled by the topology encoded in $Q$.

This viewpoint connects naturally to prior work showing that interacting copies can improve optimization. Replicated simulated annealing and SGD \cite{baldassi2016unreasonable}, Elastic SGD \cite{zhang2015deep}, and Entropy-SGD \cite{chaudhari2019entropy} all smooth the effective loss landscape through interactions among replicas, coupling forces, or local averaging in parameter space. Unlike standard replica-based optimization methods, the \(M\)-cover construction originates from graph-cover formulations of the Bethe approximation~\cite{vontobel2013counting,lucibello2014finite,altieri2017loop,leleu2026reshaping}. Our framework extends these approaches by making the topology of inter-replica communication itself a controllable degree of freedom. Different choices of $Q$ generate qualitatively different interaction structures: replicas may communicate only locally, interact globally, or exchange information directionally through the system.

We validate the effectiveness of the proposed graph-cover-based learning framework across a broad range of models and optimization algorithms, from simple perceptron learning with simulated annealing to committee machines and multilayer neural networks trained with stochastic gradient descent (SGD). These experiments demonstrate that the method is not tied to a particular architecture or learning rule, but instead provides a general mechanism for improving optimization through structured cross-replica interactions. We numerically compare our approach against existing replica-coupling methods and show that the proposed graph-cover construction consistently achieves competitive or superior performance across multiple settings.

\section{Related work}

\paragraph{Graph lifts and Bethe approximations.}
The $M$-layer construction originates from statistical physics and graphical models, where graph covers provide a controlled interpolation between the original model and its Bethe approximation \cite{vontobel2013counting,lucibello2014finite,altieri2017loop}.  In this framework, replicating the graph and rewiring edges dilutes short loops while preserving local interactions.  Our work leverages this idea as an optimization tool, extending it with a structured mixing kernel $Q$ to control how errors propagate across layers. In the following, we use the term \emph{cover} rather than \emph{layer} to avoid confusion with neural-network layers.

\paragraph{Replica-based optimization.}
Maintaining multiple interacting copies of a model has been explored in several contexts.  Replicated simulated annealing and robust ensembles improve optimization in discrete networks by encouraging exploration of dense solution regions \cite{baldassi2016unreasonable,angelini2019monte}.  In continuous settings, Elastic SGD and replicated SGD couple workers through attractive interactions \cite{zhang2015deep,pittorino2021entropic}, while Entropy-SGD optimizes a local-entropy objective favoring flat minima \cite{chaudhari2019entropy}.  These approaches primarily act in parameter space. In contrast, the $M$-cover transform acts on the routing of interactions, preserving local losses while reshaping global error-flow loops.

\section{Method}
\label{sec:method}

\subsection{General framework}

We extend the \(M\)-cover construction of~\cite{leleu2026reshaping} to supervised learning.  The starting point is to represent the training loss as a factor graph: the variable nodes are the trainable parameters, and the factor nodes are the local interactions between these parameters induced by the data and the
loss.  In the neural-network examples below, these variable nodes are weights.  Thus, the routed construction should be understood as acting on messages between trainable parameter copies: source weights contribute to local computations whose resulting error signals are assigned to destination weights.  We first present the construction in this general factor-graph setting, and then specialize it to progressively more explicit examples, including perceptrons, committee machines, and multilayer neural networks.

Let the training loss of the base model be written in factorized form as
\begin{align}
L(w)
=
\sum_{a\in A}
\ell_a(w_{\partial a}),
\label{eq:base_factorized_objective}
\end{align}
where \(A\) denotes the set of local loss factors associated with the training dataset, \(a\) indexes an individual factor, and \(\partial a\) denotes the set of trainable parameters entering factor \(a\). In supervised learning, each factor typically corresponds to a local contribution induced by one training example or minibatch element. The index \(i\) labels a trainable variable; in neural networks this is typically a synaptic parameter, e.g. a pair \((k,j)\) in a committee machine or a triple \((\ell,r,j)\) in an MLP. The \(M\)-cover transform constructs an \(M\)-cover version of the learning factor graph. Each trainable variable \(w_i\) is replaced by \(M\) cover copies
\begin{align}
w_i^{(\alpha)},
\qquad
\alpha=1,\dots,M,
\end{align}
where \(\alpha\) is a cover index.

The key operation is to preserve the local interactions encoded by each factor while changing which cover copies participate in them.  To define this assignment at the level of individual trainable variables, we distinguish the destination variable receiving a local update from the source variables providing its computational context.  We therefore refine each factor into directed incidence losses \(\ell_{a\to i}\), one for each \(i\in\partial a\), satisfying
\begin{align}
\ell_a(w_{\partial a})
=
\frac{1}{|\partial a|}
\sum_{i\in\partial a}
\ell_{a\to i}
\bigl(
w_i,
\{w_j\}_{j\in\partial a\setminus i}
\bigr).
\label{eq:directed_incidence_decomposition}
\end{align}
A directed incidence \(a\to i\) should be read as the local contribution of factor \(a\) associated with the destination parameter \(w_i\).  The variables \(w_j\), \(j\neq i\), are the source parameters that form the computational context for that contribution.

For every directed incidence \(a\to i\), and for every source variable
\(j\in\partial a\setminus i\), we draw a permutation of the cover indices,
\begin{align}
\rho_{a\to i,j}:\{1,\dots,M\}\to\{1,\dots,M\}.
\end{align}
The local contribution associated with destination copy \(w_i^{(\alpha)}\) is
then evaluated on a routed context:
\begin{align}
\ell_{a\to i}^{(\alpha)}(W\mid\rho)
=
\ell_{a\to i}
\!\left(
w_i^{(\alpha)},
\{\,w_j^{(\rho_{a\to i,j}(\alpha))}:
j\in\partial a\setminus i\,\}
\right).
\label{eq:mlayer_routed_local_loss}
\end{align}
The routing changes only the cover copy from which each source variable is read.  It does not change the structural identity of the parameter: source \(w_j\) is still source \(w_j\), but it may be read from another cover.  The destination copy remains \(w_i^{(\alpha)}\), while the surrounding source context is assembled from routed cover copies.

The routing permutations are sampled from a structured cover kernel
\(Q\in\mathbb R_+^{M\times M}\). For a permutation \(\rho\), we use
\begin{align}
\mathbb P_Q(\rho)
=
\frac{1}{Z_Q}
\prod_{\alpha=1}^M
Q_{\alpha,\rho(\alpha)},
\qquad
Z_Q=\operatorname{perm}(Q),
\label{eq:mlayer_Q_weighted_permutation_law}
\end{align}
where \(\operatorname{perm}(Q)\) denotes the matrix permanent of \(Q\).
The matrix \(Q\) weights complete matchings between covers.  It defines the topology of routed source-to-destination interactions: a sharply diagonal \(Q\) favors nearly identity routing, a uniform \(Q\) makes all permutations equally likely, and structured choices of \(Q\) induce local, diffusive, drifted, or spatially organized transport across the cover dimension.

For a fixed routing realization, the lifted training loss is
\begin{align}
\mathcal L_M(W\mid\rho)
=
\sum_{a\in A}
\frac{1}{|\partial a|}
\sum_{i\in\partial a}
\sum_{\alpha=1}^M
\ell_{a\to i}^{(\alpha)}(W\mid\rho).
\label{eq:mlayer_quenched_objective}
\end{align}
At \(M=1\), all routes are identity and \(\mathcal L_1(w)=L(w)\).  For \(M>1\), the local factors are unchanged, but their global connectivity is different.  This is the central effect of the
\(M\)-cover: it preserves local interactions while modifying the long loops through which source-to-destination messages propagate~\cite{leleu2026reshaping}.

Training uses the same local update rule as the base model, applied to the routed objective.  For differentiable models this is ordinary backpropagation through the routed computation.  For discrete models it may be a finite-difference, Glauber, repair, or straight-through update.  In all cases, the error signal generated by a routed local computation is assigned to the parameter copy that supplied the corresponding source.  Equivalently, the backward error is transported through the inverse of the route used in the
forward evaluation.

This gives the basic interpretation of the method.  The \(M\)-cover does not change the local loss, and it does not require a new optimizer.  It changes the factor graph on which the optimizer acts.  Copies share information because each copy receives error signals computed in contexts assembled from other copies.  The kernel \(Q\) controls these contexts and therefore controls the global error-feedback loops of learning.

\begin{figure*}[t]
    \centering

    \begin{subfigure}[t]{0.48\textwidth}
        \centering
        \includegraphics[width=\linewidth]{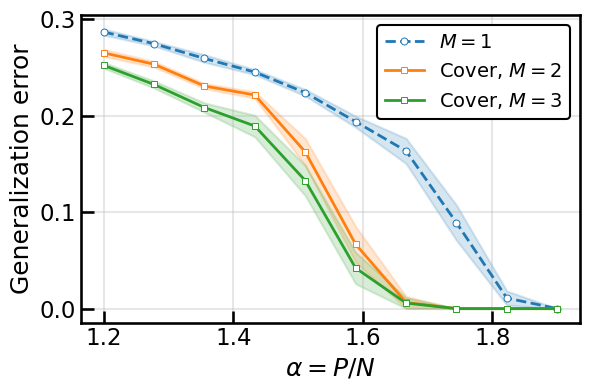}
        \caption{Perceptron}
        \label{fig:perceptron_a}
    \end{subfigure}
    \hfill
    \begin{subfigure}[t]{0.48\textwidth}
        \centering
        \includegraphics[width=\linewidth]{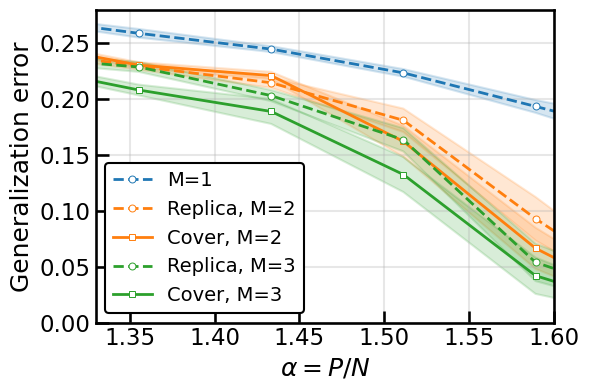}
        \caption{Perceptron}
        \label{fig:perceptron_b}
    \end{subfigure}

    \vspace{0.5cm}

    \begin{subfigure}[t]{0.48\textwidth}
        \centering
        \includegraphics[width=\linewidth]{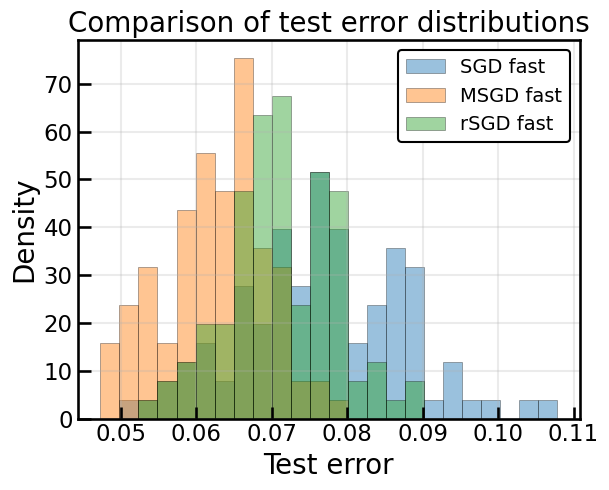}
        \caption{Committee machine}
        \label{fig:committee}
    \end{subfigure}
    \hfill
    \begin{subfigure}[t]{0.48\textwidth}
        \centering
        \includegraphics[width=\linewidth]{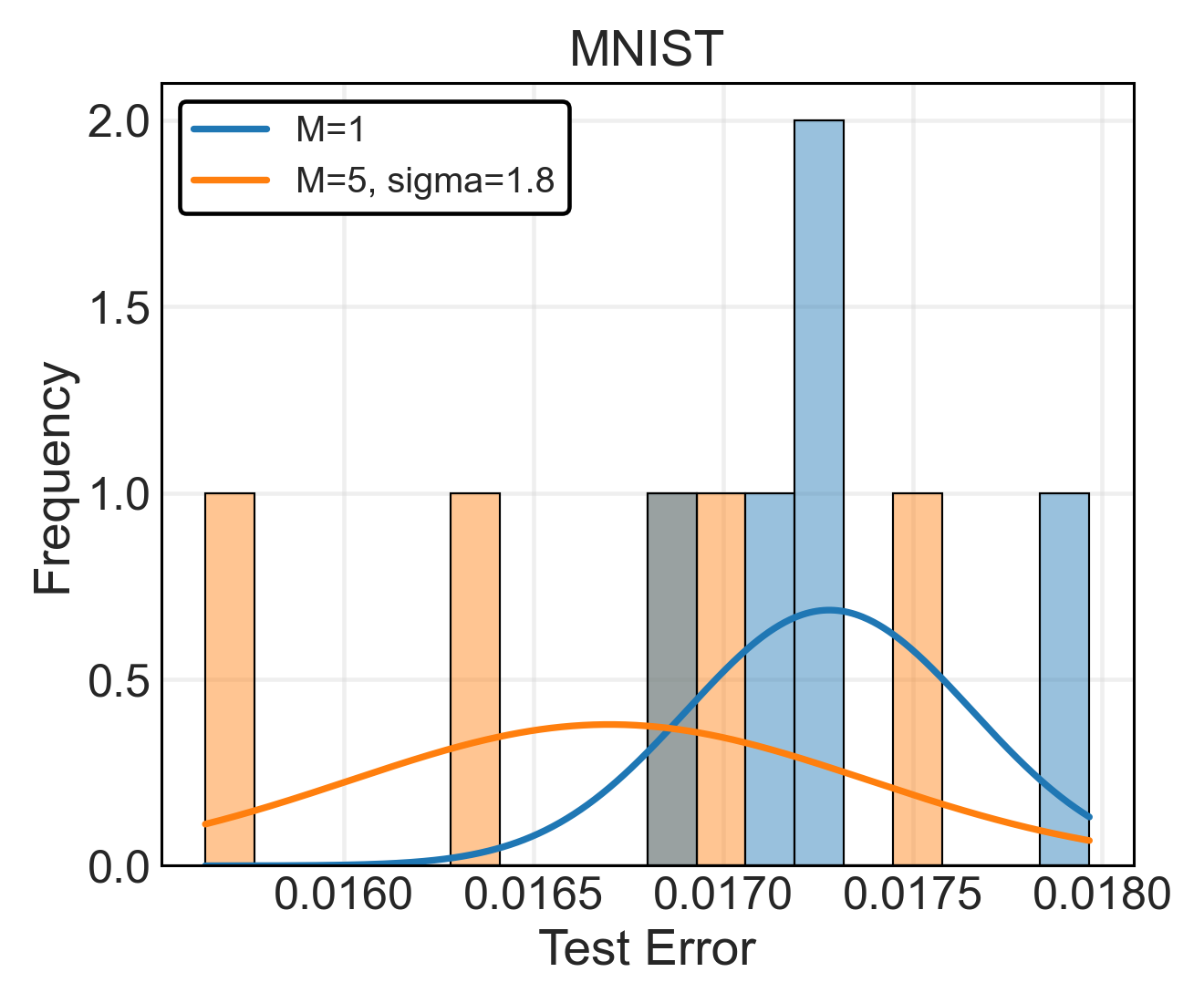}
        \caption{MLP}
        \label{fig:mlp}
    \end{subfigure}

    \caption{
    Generality of the \(M\)-cover transform across learning architectures. (a,b) Generalization error of binary teacher--student perceptrons as a function of the loading factor \(\alpha=P/N\), where \(P\) is the number of training
    patterns and \(N\) is the input dimension. (c) Test-error distribution for committee machines trained on subsampled
    Fashion-MNIST using SGD, \(M\)-cover SGD, and replicated SGD.
    (d) Test-error distribution for MNIST MLPs comparing the single-cover baseline (\(M=1\)) with structured \(M\)-cover training at \(M=5\). Across all four settings, structured error routing improves or shifts performance relative to the corresponding baselines.
    }
    \label{fig:four_architectures}
\end{figure*}

\begin{figure*}[t]
    \centering
    \includegraphics[width=1.0\textwidth]{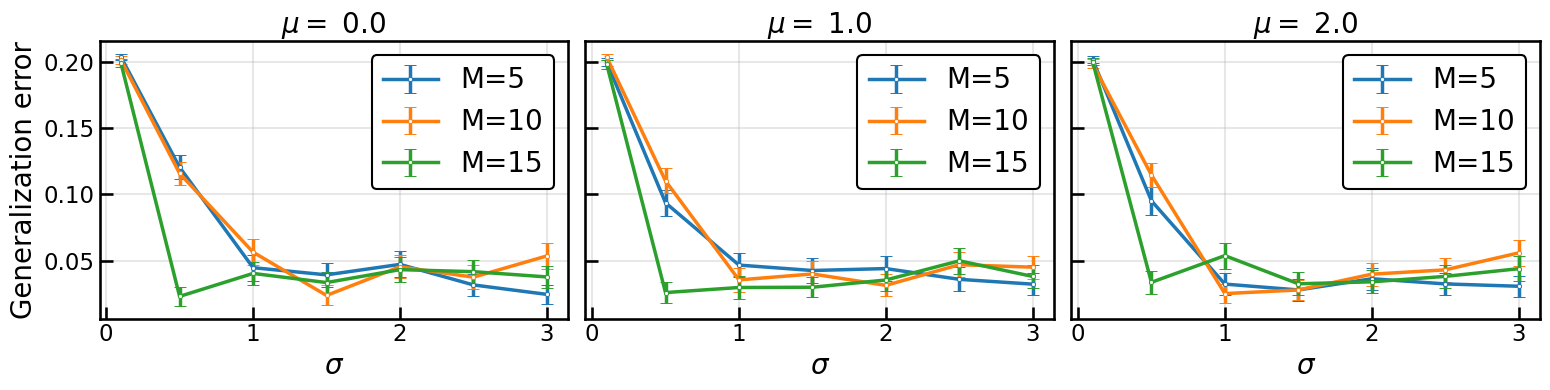}
    \caption{
        Effect of the routing kernel \(Q\) in the binary teacher--student perceptron.
        Generalization error is plotted against the Gaussian-ring width \(\sigma\) for
        different numbers of covers \(M\). Panels show increasing routing shifts
        \(\mu\). Experiments use \(N=1000\) and fixed loading \(\alpha=P/N\). Error
        bars denote 95\% CI over independent runs.
    }
    \label{fig:perceptron_sweep}
\end{figure*}

\subsection{Illustration on simple models}

\paragraph{Perceptron.}
For a teacher--student perceptron, the trainable variables are the weights \(w_q\), where \(q\) indexes a weight.  Each training pattern \(\mu\) defines one dense interaction factor through the margin
\begin{align}
\Delta_\mu
=
\frac{1}{\sqrt N}
\sum_q G_{\mu q} w_q ,
\end{align}
where \(G_{\mu q}\) is the label-gauged input pattern matrix defined in Appendix~\ref{app:perceptron}. To construct the \(M\)-cover, we choose a destination weight \(p\) and route the source weights \(q\) contributing to its local computation.  The routed margin \(\Delta_{\mu\to p}^{(\alpha)}\) is therefore obtained by replacing each source weight \(w_q\) by its routed cover copy \(w_q^{(\rho_{q\to p}(\alpha))}\), while keeping the destination weight in cover \(\alpha\).

This introduces one routing permutation \(\rho_{q\to p}\) for each ordered pair of weights \(q\to p\), leading to \(O(N^2)\) routed incidences instead of \(O(N)\) in the single-copy model. Consequently, the lifted system stores \(O(MN^2)\) routed connectivity
information and \(O(MNP)\) routed margins, while a local weight update affects \(O(NP)\) routed cavity margins rather than a single global margin.

In the exact directed lift, every destination parameter receives its own routed cavity contribution, leading to \(O(N^2)\) routed incidences in the perceptron case.  The implementation therefore introduces a \emph{sampled destination-bank} approximation: only \(S\ll N\) destination incidences are explicitly retained, while each retained routed cavity interaction remains exact.  This reduces the update complexity from \(O(PN)\) to \(O(PS)\) per weight update.

\paragraph{Committee machine.}
For a committee machine, the trainable variables are the weights
\(J_q\), where \(q=(k,i)\) indexes the weight from input coordinate \(i\) to hidden unit \(k\).  Each pattern \(\mu\) first forms hidden fields
\begin{align}
F_{\mu k}
=
\sum_{q=(k,i)}
x_{\mu i} J_q ,
\end{align}
which are then passed through a nonlinear vote or output rule.  To apply the \(M\)-cover, we choose a destination weight \(p\) and route the source weights \(q\) entering the local computation associated with \(p\).  The routed hidden field is
\begin{align}
F_{\mu k\to p}^{(\alpha)}
=
\sum_{q=(k,i)}
x_{\mu i}
J_q^{(\rho_{q\to p}(\alpha))},
\qquad
\rho_{p\to p}=\mathrm{id}.
\end{align}
Thus the structural weight \(q=(k,i)\) is unchanged, but its cover copy is routed according to the source--destination pair \(q\to p\).  Compared with the perceptron, the new feature is that routing now acts before an intermediate nonlinearity: the hidden representation itself is assembled from routed synaptic contexts before the committee output is computed.

In the exact directed lift, the destination \(p\) ranges over all weights of the committee machine.  If \(N_{\rm syn}=Kn\) is the number of weights, this requires one source-to-destination routing \(\rho_{q\to p}\) for each ordered pair of weights, hence \(O(N_{\rm syn}^2)\) routed incidences.  Storing all directed hidden fields \(F_{\mu k\to p}^{(\alpha)}\) requires \(O(MPN_{\rm syn}K)\) memory, and a local update of one weight copy affects the routed fields associated with all destination weights, giving an \(O(PN_{\rm syn}K)\) update cost if committee outputs are recomputed
explicitly.  As in the perceptron case, this cost can be reduced by sampling a destination bank \(\{p_s\}_{s=1}^S\).  One then stores only
\[
\widetilde F_{\mu s k}^{(\alpha)}
=
F_{\mu k\to p_s}^{(\alpha)},
\]
so the memory becomes \(O(SMPK)\) for the routed hidden fields and
\(O(SMN_{\rm syn})\) for the routing bank, while a local update costs
\(O(SPK)\).  The retained fields are exact directed \(M\)-cover fields; only the average over destination weights is replaced by a Monte Carlo average. Taking \(S=N_{\rm syn}\) recovers the full directed lift.

\paragraph{Multilayer perceptron.}
For an MLP, the trainable variables are synaptic weights
\(W_q\), where \(q=(\ell,r,i)\) indexes the incoming weight from unit \(i\) to neuron \(u=(\ell,r)\).  The local aggregation associated with neuron \(u\) is the preactivation
\begin{align}
a_{\mu u}
=
\sum_{q=(u,i)}
W_q h_{\mu i}.
\end{align}
In contrast to the perceptron and first-layer committee machine, the source \(h_{\mu i}\) is not fixed data: it is itself a message produced by previous layers.  The coherent \(M\)-cover therefore routes the whole source incidence \(W_q h_{\mu i}\).  For a destination weight \(p\), the lifted preactivation is
\begin{align}
a_{\mu,u\to p}^{(\alpha)}
=
\sum_{q=(u,i)}
W_q^{(\rho_{q\to p}(\alpha))}
h_{\mu i}^{(\rho_{q\to p}(\alpha))}.
\end{align}
Thus the weight and the activation it consumes are read from the same cover. This is the factor-graph lift of an MLP affine factor.

The fully directed construction is expensive because each affine factor contains many incoming weights.  For a layer with input width \(d_{\ell-1}\) and output width \(d_\ell\), a weight-level directed lift would require, for each output neuron, source-to-destination routes between incoming weights, giving \(O(d_\ell d_{\ell-1}^2)\) routed incidences for that layer.  The implementation therefore uses a \emph{block-message} approximation.  The incoming coordinates are partitioned into blocks \(g\), and the coherent source contribution of each block is first computed inside each cover,
\begin{align}
C_{\ell r g}^{\mu,(\alpha)}
=
\sum_{i\in g}
W_{\ell r i}^{(\alpha)}
h_{\ell-1,i}^{\mu,(\alpha)}.
\end{align}
The sampled routes are then collapsed into empirical mixing matrices
\(\widehat Q_{\ell r g}\), and the routed preactivation is obtained by mixing these block messages across covers,
\begin{align}
a_{\ell r}^{\mu,(\beta)}
=
\sum_g
\sum_{\alpha=1}^M
\widehat Q_{\ell r g,\beta\alpha}
C_{\ell r g}^{\mu,(\alpha)}.
\end{align}
This keeps only \(M\) activation streams rather than \(SM\) routed streams and avoids explicitly storing all source--destination weight routes.  The code implements this collapsed block-message strategy by building cached empirical mixers \(\widehat Q\) from sampled permutation banks before training, then using batched GPU operations to compute and mix the block contributions.  The exact \(Q\)-mixing mode replaces the empirical \(\widehat Q\) by the deterministic
kernel \(Q\).

\subsection{Efficient implementation}
\label{sec:subsampling_channels_cost}

The examples above are summarized in Table~\ref{tab:mlayer_examples}. Each row identifies the local aggregation used by the base model and the corresponding exact \(M\)-cover aggregation. The purpose is to show that the same operation is repeated across architectures: source trainable variables are read from routed covers, while the local
computation itself is unchanged.

The exact directed \(M\)-cover is often too expensive to instantiate fully. Indeed, if every destination parameter receives its own routed cavity contribution, the number of directed source--destination incidences grows quadratically in the number of trainable variables participating in a local factor.  In practice, we therefore use sampled destination banks or routed channels.  These approximations retain exact routed local computations on the sampled incidences, but replace the full destination average by a finite sample.

Table~\ref{tab:mlayer_costs} summarizes the resulting scaling.  Here \(M\) is the number of covers, \(S\) is the number of sampled destinations or routed channels, \(P\) is the number of data points or patterns, and \(N_{\rm syn}\) denotes the number of trainable weights.  For the MLP row, \(d_{\ell-1}\) and \(d_\ell\) are the input and output widths of layer \(\ell\), \(B\) is the minibatch size, and \(G_\ell\) is the number of input blocks used in the block-message approximation.

\begin{table*}
\centering
\small
\setlength{\tabcolsep}{6pt}
\renewcommand{\arraystretch}{1.35}

\begin{tabular}{p{2.0cm} p{2.8cm} p{4.3cm} p{6.0cm}}
\toprule

\textbf{Model}
&
\textbf{Index convention}
&
\textbf{Base local interaction}
&
\textbf{Exact \(M\)-cover lifted interaction}
\\

\midrule

Perceptron
&
\(q,p\) index weights.\newline
\(p\): destination weight.\newline
\(q\): source weight.
&
\begin{align*}
\Delta_\mu
=
\frac{1}{\sqrt N}
\sum_q G_{\mu q} w_q
\end{align*}
&
\begin{align*}
\Delta_{\mu\to p}^{(\alpha)}
=
\frac{1}{\sqrt N}
\sum_q
G_{\mu q}
w_q^{(\rho_{q\to p}(\alpha))}
\end{align*}
\\

\midrule

Committee machine
&
\(q=(k,i)\) indexes a weight into hidden unit \(k\).\newline
\(p\): destination weight.
&
\begin{align*}
F_{\mu k}
=
\sum_{q=(k,i)}
x_{\mu i} J_q
\end{align*}
&
\begin{align*}
F_{\mu k\to p}^{(\alpha)}
=
\sum_{q=(k,i)}
x_{\mu i}
J_q^{(\rho_{q\to p}(\alpha))}
\end{align*}
\\

\midrule

MLP
&
\(u=(\ell,r)\) indexes a neuron.  \newline
\(q=(u,i)\): incoming weight to \(u\).  \newline
\(p\): destination weight.
&
\begin{align*}
a_{\mu u}
=
\sum_{q=(u,i)}
W_q h_{\mu i}
\end{align*}
&
\begin{align*}
a_{\mu,u\to p}^{(\alpha)}
=
\sum_{q=(u,i)}
W_q^{(\rho_{q\to p}(\alpha))}
h_{\mu i}^{(\rho_{q\to p}(\alpha))}
\end{align*}
\\

\bottomrule
\end{tabular}

\caption{
Examples of the exact \(M\)-cover construction applied to progressively richer architectures.  The index \(q\) denotes a source trainable variable and \(p\) denotes a destination trainable variable.  The permutation \(\rho_{q\to p}\) changes only the cover index of the source variable; it does not change its structural identity.  When \(q=p\), the destination copy is kept fixed, i.e. \(\rho_{p\to p}=\mathrm{id}\).  The data-pattern index \(\mu\) is
written as a subscript throughout.  Bias terms in the MLP row are omitted for
clarity.
}
\label{tab:mlayer_examples}

\end{table*}

\begin{table*}[t]
\centering
\footnotesize
\setlength{\tabcolsep}{8pt}
\renewcommand{\arraystretch}{1.35}

\begin{tabular}{lccc}
\toprule
\textbf{Model}
&
\textbf{Base}
&
\textbf{Exact \(M\)-cover}
&
\textbf{Sampled / practical}
\\
\midrule

Perceptron
&
\(O(P)\)
&
\(O(PN)\)
&
\(O(PS)\)
\\

\midrule

Committee machine
&
\(O(PK)\)
&
\(O(PN_{\rm syn}K)\)
&
\(O(PSK)\)
\\

\midrule

MLP, layer \(\ell\)
&
\(O(Bd_{\ell-1}d_\ell)\)
&
\(O(MB d_\ell d_{\ell-1}^2)\)
&
\(O(MB d_{\ell-1}d_\ell + M^2 B d_\ell G_\ell)\)
\\

\bottomrule
\end{tabular}

\caption{
Update or forward--backward computational scaling for the base models, exact directed \(M\)-cover lifts, and sampled or block/channel implementations.  Here \(M\) is the number of covers, \(S\) the number of sampled destinations or routed channels, \(P\) the number of patterns, \(K\) the number of hidden units, \(N_{\rm syn}\) the number of trainable weights, \(B\) the minibatch size, and \(G_\ell\) the number of blocks in layer \(\ell\).  When \(S\) or \(G_\ell\) is a small constant, e.g. fewer than ten, the practical cost is close to \(M\) times the base computation up to a small constant prefactor.
}
\label{tab:mlayer_costs}
\end{table*}


\section{Results}

\subsection{Generalization gains}

To test the \(M\)-cover transform, we apply it to the three models used as illustrations in the method section. The recipe is the same in all cases: (1) start from the original learning problem, including the data, network architecture, learning dynamics, and hyperparameters; (2) apply the \(M\)-cover transform, which in practice amounts to rerouting the relevant messages as described in the method section and illustrated in Figure~\ref{fig:mlayer_workflow}; (3) run the same learning dynamics on the lifted graph, using the same number of steps or epochs; and (4) read out
the test error by collapsing the \(M\) copies of the weights into a single network and evaluating this network on held-out test data in the usual way. We stress that the final network used for evaluation has the same size as the original model. The factor of \(M\) increase in model copies is used only during learning.

Figure~\ref{fig:four_architectures} summarizes the resulting behavior across three increasingly complex architectures. For the binary teacher--student perceptron, the \(M\)-cover transform reduces generalization error over a range of loading factors \(\alpha=P/N\), where \(P\) is the number of training patterns and \(N\) is the input dimension. In this setting, increasing \(M\) improves generalization, and the gain is larger than that obtained by the replicated simulated annealing baseline~\cite{catania2024copycat}. For committee machines trained on
subsampled Fashion-MNIST, the \(M\)-cover version of SGD shifts the distribution of test errors toward lower values compared with both standard SGD and replicated SGD \cite{pittorino2021entropic}. Finally, for multilayer perceptrons trained on MNIST, structured \(M\)-cover training with \(M=5\) improves over the single-cover baseline \(M=1\). These results show that the same routing-based construction applies across discrete, shallow, and fully differentiable neural-network models.


\subsection{Dependence on mixing topology}

One advantage of the \(M\)-cover lift is that the topology of cover-to-cover interactions can be controlled explicitly through the mixing kernel \(Q\). To illustrate this dependence, we consider a one-dimensional ring structure in cover space. The kernel \(Q\) is taken to be a Gaussian ring kernel with width \(\sigma\) and shift \(\mu\). This choice is reminiscent of tail-biting spatially coupled low-density parity-check (LDPC) topologies~\cite{kudekar2011threshold}, where information propagates locally along a spatial direction.

Figure~\ref{fig:perceptron_sweep} shows that, in the binary teacher--student perceptron, the generalization error depends nontrivially on both \(\sigma\) and \(\mu\). The best performance is obtained at finite values of these parameters rather than in the nearly diagonal or fully mixed limits. This suggests that allowing error signals to propagate spatially along the cover dimension can improve generalization. The effect is reminiscent of the role of spatial structure in previous applications of the \(M\)-cover construction to
spin glasses and combinatorial optimization~\cite{leleu2026reshaping}, and more broadly of spatially coupled LDPC codes~\cite{kudekar2011threshold}.

\begin{figure}[t]
    \centering
    \includegraphics[width=0.45\textwidth]{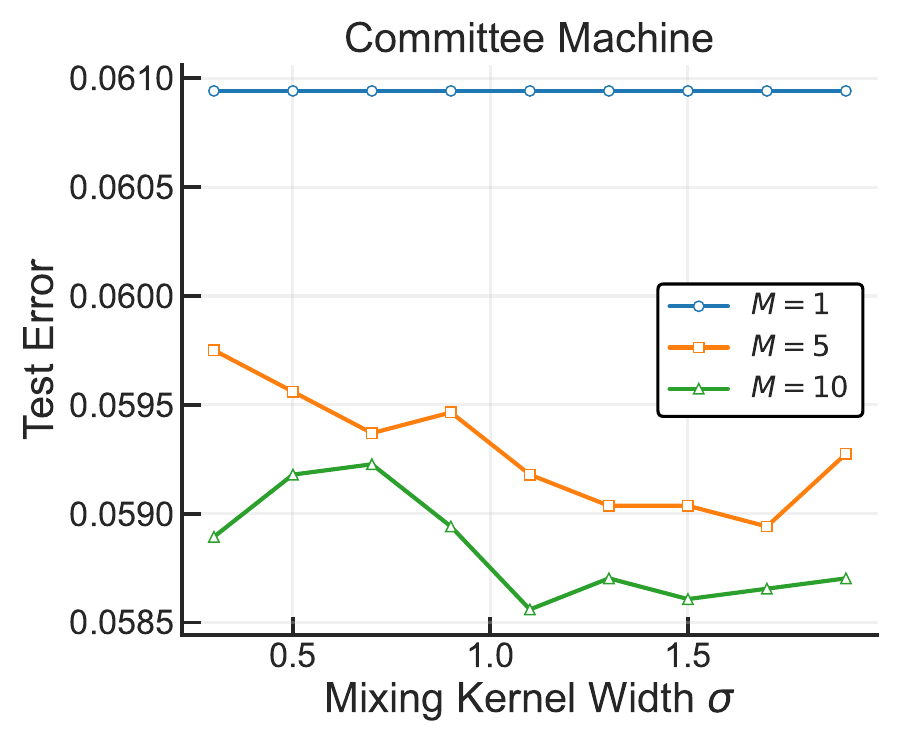}
    \caption{
        Effect of the routing kernel \(Q\) in committee machines trained on
        subsampled Fashion-MNIST. Test error is plotted against the Gaussian-ring
        mixing width \(\sigma\) for different numbers of covers \(M\). Structured
        \(M\)-cover routing improves over the single-cover baseline, with larger
        \(M\) giving lower test error across a broad range of \(\sigma\).
    }
    \label{fig:committee_sample}
\end{figure}

Figure~\ref{fig:committee_sample} shows that the same qualitative dependence appears in a more structured model: committee machines trained on subsampled Fashion-MNIST. Again, finite-width structured mixing improves over the single-cover baseline across a broad range of \(\sigma\). Larger-scale sweeps, including for multilayer perceptrons, are more computationally expensive and are left for future work. However, the optimal region observed here is
consistent across architectures, with the best performance occurring for \(\mu\) and \(\sigma\) of order one.

\begin{figure}[t]
    \centering
    \includegraphics[width=0.45\textwidth]{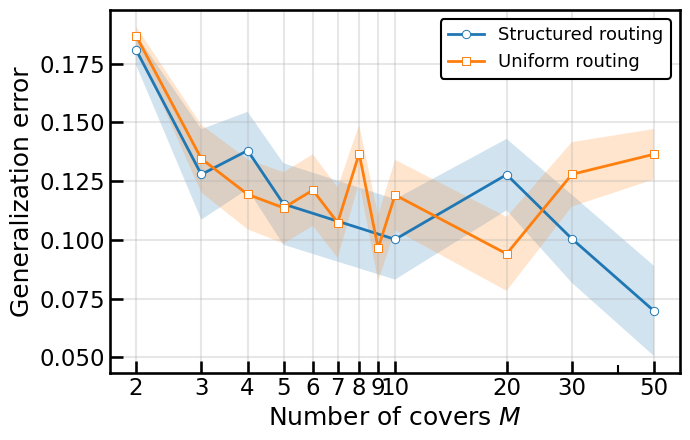}
    \caption{
Effect of the number of covers \(M\) in the binary teacher--student perceptron. 
The blue curve shows structured routing based on a Gaussian ring kernel \(Q\), parameterized by \(\mu\) and \(\sigma\), while the orange curve shows uniform routing, corresponding to the \(\sigma \to \infty\) limit. 
Shaded regions indicate the standard error across independent trials.
    }
    \label{fig:M_structured_uniform}
\end{figure}

Finally, Figure~\ref{fig:M_structured_uniform} compares structured and uniform routing as the number of covers \(M\) is varied. In the binary perceptron, structured routing continues to reduce the generalization error as \(M\) increases, whereas uniform routing gives a weaker and less systematic dependence on \(M\). This indicates that the topology of the routing kernel is not merely an implementation detail: structured cover-to-cover communication
can access generalization regimes that are harder to reach with unstructured replica methods such as replicated SGD.



\subsection{Benchmark}

Table~\ref{tab:main_results} summarizes the main quantitative comparison across the three architectures studied above. We compare the single-copy baseline, the corresponding replica-based baseline when available, and the \(M\)-cover method. Across the binary perceptron, committee machine, and MLP experiments, the \(M\)-cover construction gives the lowest test or generalization error.

This benchmark is intended as a controlled proof of principle rather than an exhaustive evaluation. The goal here is to test whether the same routing-based lift can produce consistent gains across qualitatively different learning models. Larger-scale benchmarks, broader architecture classes, and more extensive hyperparameter sweeps are left for future work.


\begin{table*}[t]
\centering
\caption{
Generalization error (mean $\pm$ CI) across different architectures and optimization methods. The left column corresponds to the perceptron problem and reports the generalization error at $\alpha = 1.58$. Confidence intervals are computed over 20 trials. For the graph-cover method, we use structured routing with $\mu = 2.0$ and $\sigma = 3.0$. For replicated simulated annealing, we use $\gamma = 1.0$. The number of replicas is fixed to $M=3$. The middle column corresponds to the committee machine results. For the graph-cover method, we use uniform permutations with $M=4$ replicas.
}
\label{tab:main_results}
\renewcommand{\arraystretch}{1.2}
\setlength{\tabcolsep}{10pt}

\begin{tabular}{lccc}
\hline
Method 
& Perceptron 
& Committee Machine 
& MLP \\
\hline

Vanilla
& $0.1936 \pm 0.0113$
& $0.076 \pm 0.002$
& $0.0173 \pm 0.001$ \\

RSA / rSGD \cite{pittorino2021entropic}
& $0.0539 \pm 0.0329$
& $0.071 \pm 0.001$
& N/A \\

M-cover
& $\mathbf{0.0420 \pm 0.0319}$
& $\mathbf{0.062 \pm 0.001}$
& $\mathbf{0.0167 \pm 0.001}$ \\

\hline
\end{tabular}

\end{table*}

\section{Conclusion}

We introduced an \(M\)-cover extension of machine learning models based on a simple recipe: replicate the model, reroute the interactions between copies, run the original learning dynamics on the lifted graph, and collapse the copies at readout. Across binary teacher--student perceptrons, committee machines, and multilayer perceptrons, this construction improves generalization relative to the corresponding single-copy baselines and, in the settings tested here, also improves over classical replicated methods such as
replicated simulated annealing and replicated SGD.

A central appeal of the approach is its generality. The method does not require designing new interaction potentials, adding explicit attractive forces between replicas, annealing additional parameters, or modifying the local optimizer. Instead, the same local learning rule is applied on a graph whose global connectivity has been changed by the \(M\)-cover routing. In the uniform-mixing limit, the method has essentially no topology-specific hyperparameters beyond
the number of covers, yet already gives competitive performance. Moreover, the computational cost can be controlled by sampling destination channels or permutation banks, suggesting a practical route toward scaling the method to larger architectures.

The results also indicate that the topology of the cover is itself a useful degree of freedom. Structured kernels \(Q\), such as local Gaussian ring couplings, can outperform uniform routing and can continue improving with the number of covers. This connects the present learning framework to the broader role of spatial coupling in message-passing systems, including spatially coupled LDPC codes and recent \(M\)-cover constructions for hard optimization problems~\cite{leleu2026reshaping}. Another direction is to clarify the relationship between \(M\)-cover routing and other stochastic or conditional-computation mechanisms. Dropout can be interpreted as training an ensemble of randomly thinned subnetworks and using an averaged predictor at test time~\cite{srivastava2014dropout}, while mixture-of-experts models route examples through different expert subnetworks using gating mechanisms~\cite{jacobs1991adaptive,shazeer2017outrageously}. The \(M\)-cover construction differs from both: it routes learning messages across replicated copies of the same model and collapses these copies at readout. Nevertheless, understanding whether structured cover routing induces a form of regularization related to dropout, or whether covers can be interpreted as interacting experts, is an interesting direction for future work. A natural next step is therefore to search for routing topologies adapted to specific architectures and data distributions, and to test whether structured error transport can provide systematic gains in larger-scale learning systems.


\bibliographystyle{icml2026_fogen}  
\bibliography{references}  

\appendix

\section{Numerical experiments}

\subsection{Perceptron\label{app:perceptron}}

We evaluated the proposed \(M\)-cover routing method on the binary teacher--student perceptron problem with simulated annealing dynamics. The perceptron dimension was fixed to
\(
N=1000
\),
and the number of training patterns was
\(
P=\alpha N,
\)
with
\(
\alpha
\)
swept uniformly between \(1.2\) and \(1.9\). For each setting, we performed \(20\) independent trials with different random seeds.

The teacher--student data were generated as follows.  A teacher weight vector
\(w^\star\in\{-1,+1\}^N\) was sampled uniformly at random, and raw input
patterns \(x_{\mu q}\in\{-1,+1\}\) were sampled independently.  The label of
pattern \(\mu\) was generated by the teacher perceptron,
\begin{align}
y_\mu
=
\operatorname{sign}
\left(
\frac{1}{\sqrt N}
\sum_{q=1}^N x_{\mu q} w_q^\star
\right).
\end{align}
We then use the label-gauged pattern matrix
\begin{align}
G_{\mu q}
=
y_\mu x_{\mu q},
\end{align}
so that the student margin can be written as
\begin{align}
\Delta_\mu(w)
=
\frac{1}{\sqrt N}
\sum_{q=1}^N G_{\mu q} w_q .
\end{align}
With this convention, correct classification corresponds to
\(\Delta_\mu(w)>0\).

The proposed \(M\)-cover method used site-dependent routing permutations sampled from a directional Gaussian-ring mixing kernel \(Q\). Unless otherwise stated, experiments used
\(
\sigma=1.5
\)
and
\(
\mu=2.0.
\)
The number of sampled routing permutations was fixed to
\(
S_{\rm perm}=10.
\)
The routing matrix was balanced using Sinkhorn normalization before permutation sampling.

We compared the method against replicated simulated annealing (RSA) with coupling strength
\(
\gamma=1.0.
\)

All methods were trained using Glauber simulated annealing with single-spin updates. The annealing schedule linearly decreased the temperature from
\(
T_{\max}=0.4
\)
to
\(
T_{\min}=10^{-2}
\)
using step size
\(
10^{-4}.
\)

Generalization performance was evaluated using the standard overlap and generalization error of the teacher--student perceptron problem. For the \(M\)-cover method, the replicated covers were first collapsed into a single effective weight vector \(\bar w=\frac{1}{M}\sum_{\alpha=1}^M w^{(\alpha)}\), which was then used for evaluation. The overlap with the teacher vector \(w^\star\) was computed as \(R=\frac{\bar w\cdot w^\star}{\|\bar w\|\,\|w^\star\|}\), and the corresponding generalization error was \(\epsilon_g=\frac{1}{\pi}\arccos(R)\).

\subsection{Committee Machine}

We evaluated the proposed \(M\)-cover routing method on the binary committee machine using the Fashion-MNIST teacher--student setting introduced in prior work \cite{pittorino2021entropic}. The dataset consisted of binarized Fashion-MNIST inputs restricted to two classes, following the preprocessing procedure used in the reference implementation.

All experiments used a committee machine with
\(
K=9
\)
hidden units and mini-batch size
\(
B=100.
\)
Training was performed using stochastic gradient descent together with the same cross-entropy surrogate, annealing schedules, and optimization hyperparameters as the reference SGD implementation. In particular, the learning-rate schedule and the annealing procedure used to progressively sharpen the soft classifier into a hard classifier were kept identical across vanilla SGD, replicated SGD (rSGD), and the proposed \(M\)-cover SGD in order to ensure a fair comparison.

For the proposed \(M\)-cover method, we introduced \(M\) replicated covers together with site-dependent routing permutations. Unless otherwise stated, experiments used
\(
M=4
\)
and
\(
S_{\rm perm}=10.
\)
We used uniform routing permutations, corresponding to the limit \(\sigma\to\infty\) of the Gaussian-ring mixing kernel.

We compared the method against both vanilla SGD and replicated SGD (rSGD). For the replicated SGD baseline, both the initial coupling strength and its annealing schedule were chosen to match the ``slow SGD'' setting of the reference implementation.

All methods were trained for up to
\(
20\,000
\)
epochs. Training was stopped either when the loss fell below
\(
10^{-7}
\)
or when zero training error was reached. Generalization performance was evaluated using the hard classification error on the test set. Error bars correspond to standard errors computed across independent random trials.

\subsection{MNIST multilayer perceptron experiments}
\label{app:mlp_mnist_experiments}

We evaluated the \(M\)-cover construction on MNIST using a fully connected multilayer perceptron implemented in PyTorch.  MNIST images were flattened into vectors in \(\mathbb{R}^{784}\), rescaled to \([0,1]\), and normalized using the standard MNIST mean and standard deviation. 

The base network had architecture $784 \rightarrow 512 \rightarrow 512 \rightarrow 10$, with ReLU nonlinearities after the two hidden layers.  This corresponds to \(669{,}706\) trainable parameters per replica.  During training, all weights and biases were replicated \(M\) times, but evaluation was performed using a collapsed single-network readout obtained by averaging the replica weights.

Training used the collapsed block-message approximation described in the main text.  Coherent block messages were first computed independently inside each replica and then mixed across covers using empirical mixing matrices obtained from sampled permutation banks.  Unless otherwise
stated, routing used directional Gaussian-ring kernels \(Q\). 

Optimization was performed with SGD using momentum and Nesterov acceleration. The loss was the cross-entropy evaluated on the routed logits.

\begin{table}[t]
\centering
\small
\renewcommand{\arraystretch}{1.15}
\setlength{\tabcolsep}{6pt}
\caption{Main hyperparameters for the MNIST MLP experiments.}
\label{tab:mlp_mnist_setup}
\begin{tabular}{ll}
\toprule
Dataset & MNIST \\
Input dimension & \(784\) \\
Architecture & \(784\!-\!512\!-\!512\!-\!10\) \\
Activation & ReLU \\
Parameters per replica & \(669{,}706\) \\
Optimizer & SGD + momentum + Nesterov \\
Learning rate & \(0.05\) \\
Momentum & \(0.9\) \\
Weight decay & \(0\) \\
Batch size & \(256\) \\
Epochs & \(40\) \\
Evaluation frequency & Every epoch \\
Evaluation batch size & \(4096\) \\
Number of covers & \(M\in\{1,5,10,15\}\) \\
Routing kernel & Directional Gaussian ring \\
Permutation samples & \(S_{\rm perm}=10\) \\
Initialization & Shared initialization + Gaussian noise \\
Initialization noise std. & \(10^{-2}\) \\
Implementation & PyTorch, GPU, float32 \\
\bottomrule
\end{tabular}
\end{table}

\end{document}